\documentclass[runningheads]{llncs}
\usepackage[T1]{fontenc}
\usepackage{graphicx}
\usepackage{cite}
\usepackage{bbding}
\usepackage{amsmath}
\usepackage{amsfonts}
\usepackage{multirow}
\usepackage{graphicx}
\usepackage{booktabs}
\usepackage{xcolor}
\usepackage{amssymb}
\usepackage{pifont}
\DeclareMathOperator*{\argmin}{min}
\newcommand{\cmark}{\ding{51}}
\newcommand{\xmark}{\ding{55}}
\begin{document}

\title{Skin Lesion Recognition with Class-Hierarchy Regularized Hyperbolic Embeddings}

\author{Zhen Yu\inst{1,6,7}, Toan Nguyen\inst{5}, Yaniv Gal\inst{3}, Lie Ju\inst{6, 7},
Shekhar S. Chandra\inst{4},
Lei Zhang\inst{1},
Paul Bonnington\inst{6},
Victoria Mar\inst{2},
Zhiyong	Wang\inst{5},
Zongyuan Ge \inst{6,7, 8(}\Envelope\inst{)}
}
\index{Zhen, Yu}
\index{Lie, Ju}
\index{Victoria, Mar}
\index{Anders, Eriksson}
\index{Shekhar S. Chandra}
\index{Paul, Bonnington}
\index{Lei, Zhang}
\index{Zongyuan, Ge}

\institute{Central Clinical School, Faculty of Medicine, Nursing and Health Sciences, Monash University, Melbourne, Australia \and
Victorian Melanoma Service, Alfred Health, Melbourne, Australia \and
Kāhu, Auckland, New Zealand \and
School of Information Technology and Electrical Engineering, The University of Queensland, Brisbane Qld, Australia \and
School of Computer Science, The University of Sydney, Sydney, Australia \and
eResearch Centre, Monash University, Melbourne, Australia \and
Monash Airdoc Research, Monash University, Melbourne, Australia. \and
Faculty of Engineering, Monash University, Melbourne, Australia \\ 
\email{zongyuan.ge@monash.edu}, \url{https://mmai.group}
}

\maketitle           

\begin{abstract}
In practice, many medical datasets have an underlying taxonomy defined over the disease label space. However, existing classification algorithms for medical diagnoses often assume semantically independent labels. In this study, we aim to leverage class hierarchy with deep learning algorithms for more accurate and reliable skin lesion recognition. We propose a hyperbolic network to jointly learn image embeddings and class prototypes. The hyperbola provably provides a space for modeling hierarchical relations better than Euclidean geometry. Meanwhile, we restrict the distribution of hyperbolic prototypes with a distance matrix which is encoded from the class hierarchy. Accordingly, the learned prototypes preserve the semantic class relations in the embedding space and we can predict label of an image by assigning its feature to the nearest hyperbolic class prototype. We use an in-house skin lesion dataset which consists of  $\sim$230k dermoscopic images on 65 skin diseases to verify our method. Extensive experiments provide evidence that our model can achieve higher accuracy with less severe classification errors compared to that of models without considering class relations. 

\end{abstract}

\section{Introduction}
Recent advances in deep learning have greatly improved the accuracy of classification algorithms for medical image diagnosis. Typically, these algorithms assume mutually exclusive and semantically independent labels~\cite{yu2016automated,esteva2017dermatologist,brinker2019deep}. The classification performance is evaluated by treating all classes other than the true class as equally wrong. However, many medical datasets have an underlying class hierarchy defined over the label space. Accordingly, different disease categories can be organized from general to specific in the semantic concepts. Diseases from a same super-class often share similar clinical characteristics. Incorporating the constraint of class relations in a diagnostic algorithm has at least two benefits. First, a class hierarchy defines a prior knowledge on the structure of disease labels. Learning model with such knowledge would facilitate the model training and boost the performance compared with that of using semantic-agnostic labels. Second, a class tree indicates the semantic similarity between each class pair and a model can be optimized with the semantic metric to reduce the severity of prediction errors~\cite{bertinetto2020making, ijcai2021-337, garnot2021leveraging}. Take the example of a diagnostic model in dermatology: the common non-cancerous melanocytic lesion has at least two sub-classes: lentigo and benign nevus. Undoubtedly, mistaking a lentigo for a begin nevus is more tolerable than of mistaking a malignant melanoma for a begin nevus. By taking mistake severity into consideration, we can somewhat preclude models from making a egregious diagnostic error which is crucial in deploying the model in real world scenarios.

\begin{figure}[!t]
\centering
\includegraphics[width=0.9\textwidth]{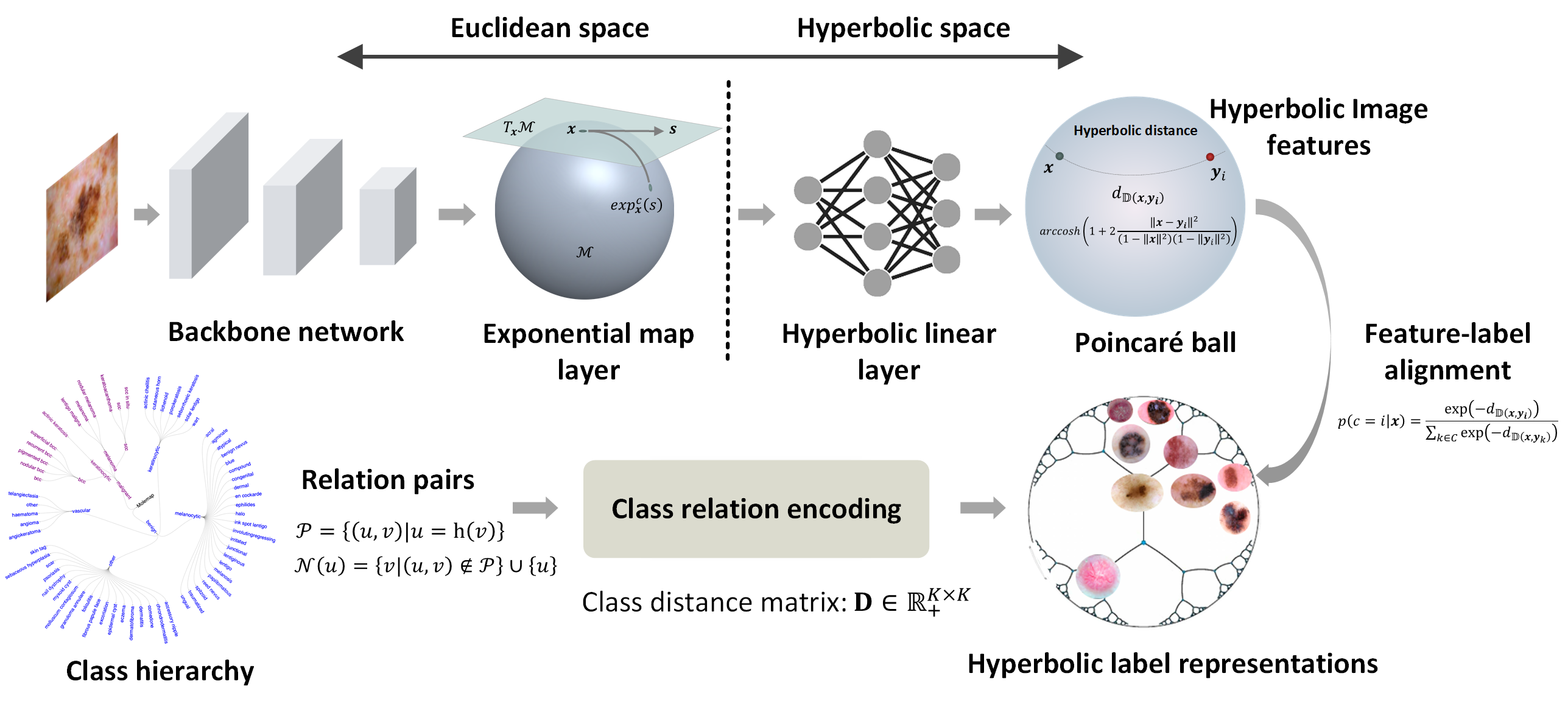}
\caption{The proposed hyperbolic model with class hierarchy for skin lesion recognition.}
\label{fig: framework}
\end{figure}

However, relatively few works use hierarchical class clues in the context of medical image analysis \cite{barata2019deep, yang2020hierarchical}. Although these methods report promising results, they require the network architecture to be adapted for a specific hierarchy and they neglect semantic measurements in evaluating the performance of the algorithms. Besides, these models are built in Euclidean space while study \cite{ganea2018hyperbolic} shows that Euclidean space suffer from heavy volume intersection and points arranged with Euclidean distances would no longer be capable of persevering the structure of the original tree. By contrast, approaches with hyperbolic geometry for modelling symbolic data have demonstrated to be more effective in representing hierarchical relations \cite{nickel2017poincare, ganea2018hyperbolic}. The hyperbolic space can reflect complex structural patterns inherent in taxonomic data with a low dimensional embedding. 

In this study, we propose modelling class dependencies in the hyperbolic space for skin lesion recognition. Our aim is to improve accuracy while reducing the severity of classification mistakes by explicitly encoding hierarchical class relations into hyperbolic embeddings. To this end, We first design a hyperbolic prototype network which is capable of jointly learning image embeddings and class prototypes in a shared hyperbolic space. Then, we guide the learning of hyperbolic prototypes with a distance matrix which is encoded from the given class hierarchy. Hence, the learned prototypes preserve the semantic class relationship in the embedding space. We can predict the label of an image by assigning its feature to the nearest hyperbolic class prototype. Our model can be easily applied to different hierarchical image datasets without complicated architecture modification. We verify our method on an in-house skin lesion image dataset which consists of approximately 230k dermoscopic images organized in three-level taxonomy of 65 skin diseases. Extensive experiments prove that our model can achieve higher classification accuracy with less severe classification errors than models without considering class relations. Moreover, we also conducted an ablation study by comparing hyperbolic space trained hierarchical models to those trained in Euclidean space. 

\section{Method}
\subsection{Hyperbolic Geometry}

The hyperbolic space $\mathbb{H}^{n}$ is a homogeneous, simply connected Riemannian manifold with constant negative curvature. There exist five insightful models of $\mathbb{H}^{n}$ and they are conformal to the Euclidean space. Following \cite{nickel2017poincare}, we use the Poincar\'e ball model because it can be easily optimized with gradient-based methods.
The Poincar\'e ball model $\left( \mathbb{D}^{n}, g^{\mathbb{D}}\right)$ is defined by the manifold $\mathbb{D}^{n} = \left\{x \in \mathbb{R}^{n}: c\left \| x \right \| < 1, c \geq 0\right\}$ endowed with the Riemannian metric $g_x^{\mathbb{D}} = \lambda_x^{2c} g^{E}$, where $c$ denotes the curvature, $\lambda_x^c=\frac{2}{1-c\left \| x \right \|^{2}}$ is the conformal factor and $g^{E}=\mathbf{I}^n$ is the Euclidean metirc tensor. The hyperbolic space has very different geometric properties than that in the Euclidean space. We introduce basic hyperbolic operations involved in this study as following:

\noindent\subsubsection{Poincar\'e Distance.} The distance between two points $\textbf{x}_1, \textbf{x}_2 \in \mathbb{D}^n_c$ is calculated as: 
\begin{equation}
\footnotesize
d_c\left(\mathbf{x}_1, \mathbf{x}_2 \right)= \frac{2}{\sqrt{c}}\text{arctanh}\left( \sqrt{c}\left\| -\textbf{x}_1 \oplus_c \textbf{x}_2 \right\| \right)
\end{equation}

\begin{equation}
\footnotesize
\textbf{x}_1 \oplus_c \textbf{x}_2 = \frac{\left( 1+2c\left\langle \textbf{x}_1, \textbf{x}_2 \right\rangle + c\left\| \textbf{x}_2 \right\|^2 \right)\textbf{x}_1 + \left( 1-c\left\| \textbf{x}_1 \right\|^2 \right)\textbf{x}_2}{1+2c\left\langle \textbf{x}_1,\textbf{x}_2 \right\rangle + c^2\left\| \textbf{x}_1 \right\|^2\left\| \textbf{x}_2 \right\|^2 }
\end{equation}
\noindent\subsubsection{Exponential map.} The \textit{exponential map} defines a projection from tangent space $T_\textbf{x}\mathbb{D}_c^{n}$ of a Riemannian manifold $\mathbb{D}_c^{n}$ to itself which enables us to map a vector in Euclidean space $\mathbb{R}^{n} \cong T_\textbf{x}\mathbb{D}_c^{n}$ to the hyperbolic manifold. The mathematical definition of \textit{exponential map} is given by:
\begin{equation}
\footnotesize
\text{exp}_\textbf{x}^{c}\left( \textbf{v} \right) = \textbf{x} \oplus_c\left( \text{tanh}\left( \sqrt{c}\frac{\lambda_\textbf{x}^{c}\left\| \textbf{v} \right\|}{2} \right) \frac{\textbf{v}}{\sqrt{c}\left\| \textbf{v} \right\|}\right)
\end{equation}
where $\mathbf{x}$ denotes the reference point and default $\mathbf{0}$ are used if not specified.
\subsubsection{Hyperbolic linear projection.} The linear projection in hyperbolic space is based on the $\text{M}\ddot{\text{o}}\text{bius}$ matrix-vector multiplication. Let $\psi: \mathbb{R}^{n} \to \mathbb{R}^{m}$ be a linear map defined in Euclidean space. Then, for $\forall \textbf{x} \in \mathbb{D}^{n}_c$, if $\psi\left( \mathbf{x}\right) \neq \textbf{0} $, the calculation of the linear map is defined as:
\begin{equation}
\footnotesize
\psi^{\otimes_c}\left( \textbf{x} \right) =  \frac{1}{\sqrt{c}} \text{tanh}\left( \frac{\left\| \psi\left( \textbf{x} \right) \right\|}{\left\| \textbf{x} \right\|} \text{tanh}^{-1}\left( \sqrt{c} \left\| \textbf{x} \right\|\right)\right)\frac{\psi\left( \textbf{x} \right)}{\left\| \psi\left( \textbf{x} \right) \right\|}
\label{eq: linear}
\end{equation}
If we consider a bias of $\textbf{b}\in\mathbb{D}^n_c$ in the linear map, then the $\mathbf{x}$ in the above equation should be replaced with the $\text{M}\ddot{\text{o}}\text{bius}$ sum: $\textbf{x}\gets \textbf{x} \oplus_c \textbf{b}$.

\subsection{Hyperbolic prototype network for image classification}
As shown in Fig. \ref{fig: framework}, the proposed hyperbolic prototype network (HPN) consists of a backbone network, a \textit{exponential map} layer, a hyperbolic linear layer and a classification layer. First, the backbone network extracts image representations from the Euclidean space and then the \textit{exponential map} layer projects it into the Poincar\'e ball. After that, we use a hyperbolic linear layer to transform the projected hyperbolic image embeddings so that their dimensions are fitted with that of class prototypes in the shared hyperbolic space. Finally, the classification layer performs matching between hyperbolic image embeddings with respect to the corresponding class prototypes.

Formally, consider a dataset $\mathcal{N}$ consists of \textit{m} samples $\left\{ \left(\textbf{x}_i, \text{y}_i \right )\right\}^{m}$ from \textit{K} classes. Let $\left\{\textbf{a}_1 , ..., \textbf{a}_K \right\}$ be the hyperbolic prototypes for the \textit{K} classes. Then, the hyperbolic network computes probability distributions over all the class prototypes for each input image as:
\begin{equation}
\footnotesize
p\left( \hat{\text{y}}_i = k|\textbf{x}_i \right) = \frac{\text{exp}\left( -d_c\left( \textbf{z}_i,\textbf{a}_k  \right) \right)}{\sum_{j=1}^{K} \text{exp}\left( -d_c\left( \textbf{z}_i, \textbf{a}_j \right) \right)}, \forall k\in K
\label{eq:dist_softmax}
\end{equation}
\begin{equation}
\footnotesize
\textbf{z}_i=\psi^{\otimes_c} \left( \text{exp}
_\textbf{0}^c\left( f\left( \textbf{x}_i \right) \right)\oplus_c\textbf{b} \right)
\end{equation}
where $f\left(\cdot\right)$ denotes the function of backbone network. The network can be directly optimized with the cross-entropy loss on the hyperbolic distance-based logits:
\begin{equation}
\footnotesize
\mathcal{L}_{DCE} = \frac{1}{m} \sum_{i \in m}^{~}\left ( \frac{1}{T}d_c\left ( \mathbf{z}_i, \mathbf{a}_{\text{y}_i} \right ) + \text{log} \left ( \sum_{j=1}^{K} \text{exp} \left ( -\frac{1}{T}d_c\left ( \mathbf{z}_i, \mathbf{a}_j \right ) \right )\right )\right )   
\label{eq: dce_loss}
\end{equation}
\noindent
where \textit{T} is a temperature factor for scaling the distance logits which is fixed as 0.1.

\begin{figure}[t!]
\centering
\includegraphics[width=\textwidth]{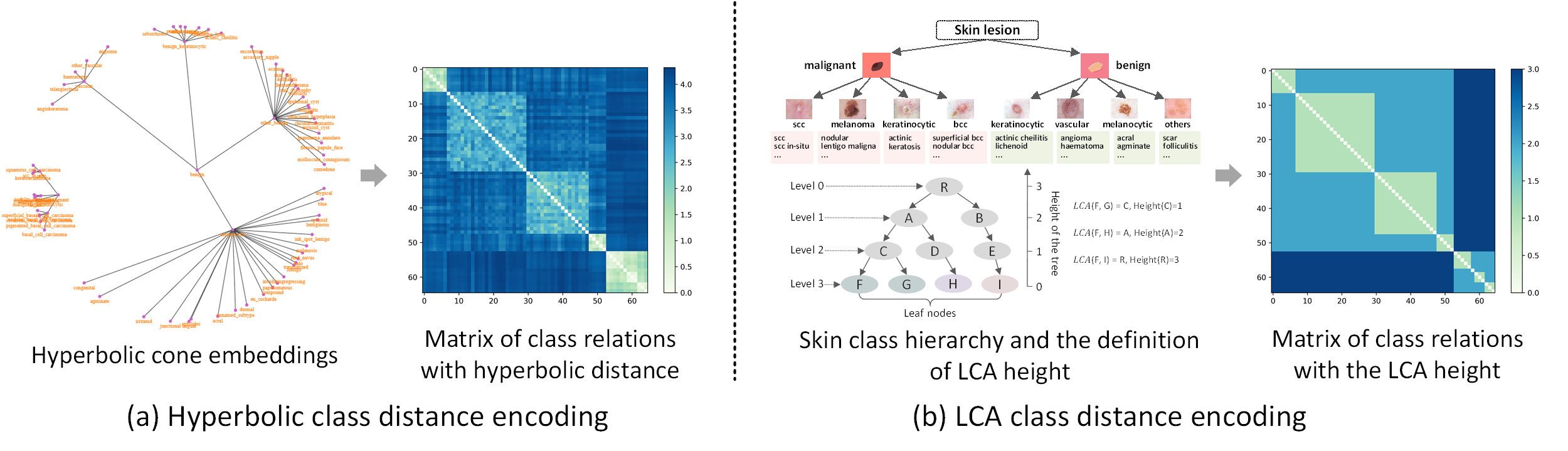}
\caption{Illustration of class relation encoding methods. Both of the approaches produce a matrix which indicates the dissimilarity of class pairs.}
\label{fig: class_distance}
\end{figure}

\subsection{Incorporating constraint of class relations.} 
\subsubsection{Hierarchical class relations encoding:} Let's assume that the label of dataset $\mathcal{N}$ can be organized into a class tree $\mathcal{H}=\left(V, E\right)$ with \textit{h} hierarchical levels, where \textit{V} and \textit{E} denote nodes and edges, respectively. Each node corresponds to a class label while an edge $\left(u, v\right) \in E$ indicates entailment relation\footnote{Namely, \textit{v} is a sub-concept of \textit{u}} between the node pair. We exploit two methods for encoding class relations from the hierarchical tree $\mathcal{H}$: 1) hyperbolic class distance (HCD) encoding, and 2) low common ancestor (LCA) class distance (LCD) encoding. Both of the methods produce a matrix of class distance $\mathbf{D}\in \mathbb{R}_+^{K \times K}$ which indicates the dissimilarity between each pair of leaf-classes (see Fig. \ref{fig: class_distance}).
For the HCD encoding, we first learn hyperbolic representations for all class labels following \cite{ganea2018hyperbolic} and then compute pairwise hyperbolic distances for leaf classes. Similar to \cite{bertinetto2020making}, in the LCD encoding, we directly define the distance between two classes as the height of their LCA in the hierarchy.   
\subsubsection{Class distance guided hyperbolic prototype learning:} 
To introduce such class relations into class prototypes, we propose to guide the prototype learning by constraining the distance between prototypes to be consistent with the class distance in the $\mathbf{D}$. As described by Sala et al. \cite{sala2018representation}, the distortion of a mapping between the finite metric space of hierarchical class distance $\mathbf{D}\left[i, j\right]$ and the continuous metric space of prototype distance $d_c\left(\mathbf{a}_i, \mathbf{a}_j\right)$ can be defined as:
\begin{equation}
\footnotesize
\text{disto}\left( \mathbf{d}, \mathbf{D}\right) = \frac{1}{K\left(K-1\right)} \sum_{i,j \in K^2,i\neq j}^{~}\frac{\left | d_c\left ( \mathbf{a}_i, \mathbf{a}_j \right ) -\mathbf{D}\left [i,j \right ] \right|}{\mathbf{D}\left [i,j \right ]}
\label{eq: disto1}
\end{equation}

\noindent
A low-distortion mapping means that the learned prototypes preserve well the relations between classes defined by the $\mathbf{D}$. However, achieving low-distortion mapping requires the prototypes to be arranged in the embedding space with the specific distance constraint, and this may conflict with the cross-entropy loss (\ref{eq: dce_loss}) which encourages the distance between an embedding to negative class prototypes to be as large as possible. Hence, we introduce a scale factor in the formulation of the distortion for removing the discrepancy between the scale of the prototype distance and the hierarchical class distance:

\begin{equation}
\footnotesize
s = \sum_{i,j \in K^2}^{}\frac{d_c\left ( \mathbf{a}_i, \mathbf{a}_j \right )} {\mathbf{D}\left [i,j \right ]}  \mathbin{/} \sum_{i,j \in K^2}^{}\frac{d_c\left ( \mathbf{a}_i, \mathbf{a}_j \right )^2} {\mathbf{D}\left [i,j \right ]^2}
\label{eq: disto2}
\end{equation}
\noindent
The \textit{s} is dynamically changed depending on the value of \textbf{d} and \textbf{D}. Then, we obtain the following smooth surrogate of the $\text{disto}^s$ for optimization:
\begin{equation}
\footnotesize
\mathcal{L}_{disto} = \frac{1}{K\left(K-1\right)} \argmin_{s\in \mathbb{R}_+} \sum_{i,j \in K^2,i\neq j}^{~} \left(\frac{s\cdot d_c\left ( \mathbf{a}_i, \mathbf{a}_j \right ) -\mathbf{D}\left [i,j \right ]}{\mathbf{D}\left [i,j \right ]}\right)^2
\end{equation}
\noindent

Finally, we optimize the proposed hyperbolic network by combining $\mathcal{L}_{DCE}$ and $\mathcal{L}_{disto}$. The $\mathcal{L}_{DCE}$ enables us to jointly learn hyperbolic image embeddings and class prototypes, while the $\mathcal{L}_{disto}$ forces the class prototypes following the semantic distribution defined by the given class hierarchy:
\begin{equation}
\footnotesize
\mathcal{L} = \mathcal{L}_{DCE} + \mathcal{L}_{disto}
\end{equation}
\noindent

 \section{Experiment and Results}
\subsection{Dataset and Implementation}
\begin{figure}[h]
\centering
\includegraphics[width=0.5\textwidth]{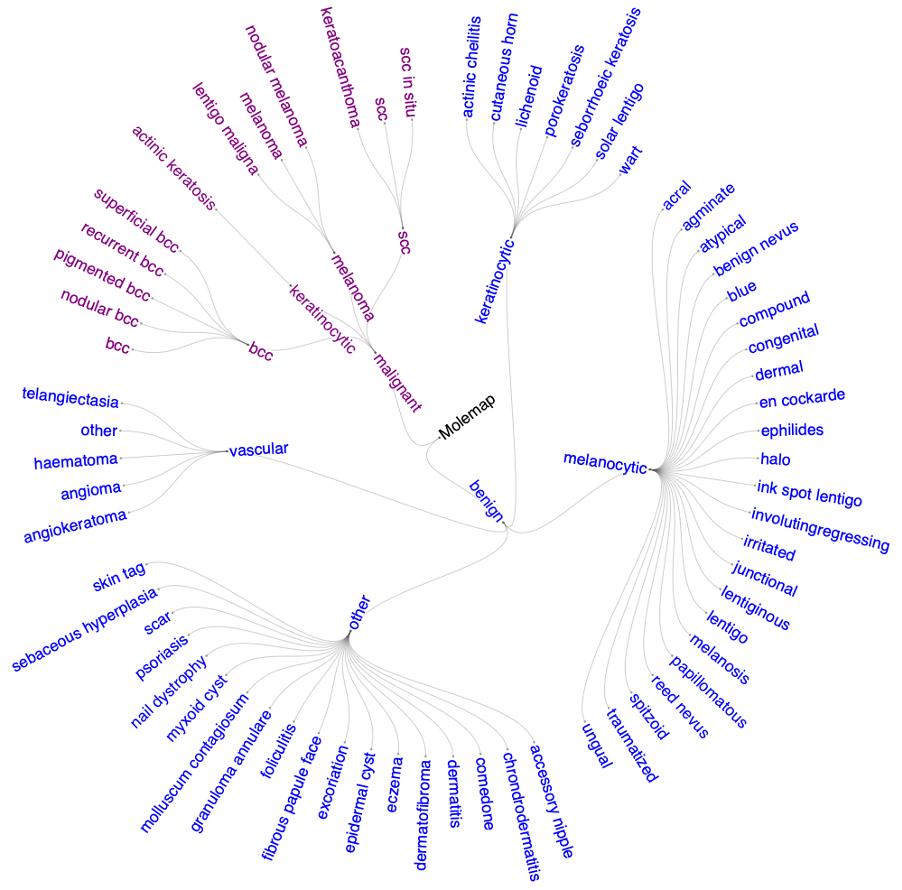}
\caption{Structure of the skin disease taxonomy.}
\label{fig: class_hierarchy}
\end{figure}

We evaluate the proposed method on an in-house dataset. We denote the dataset as $molemap^+$ as it is bigger and more diverse compared to the first version used in \cite{ge2017skin}. The $molemap^+$ includes 235,268 tele-dermatology verified dermoscopic images organized in three-level tree-structured taxonomy of 65 skin conditions (shown in Fig. \ref{fig: class_hierarchy}). We split the dataset into training, validation and testing set with a ratio of 7:1:2. The standard data augmentation techniques such as random resized cropping, colour transformation, and flipping are equally used in all experiments. Each dermoscopic image is resized to a fixed input size of 320$\times$320. We use ReseNet-34 \cite{he2016deep} as the backbone for all models and train them using ADAM optimizer with a batch size of 100 and a training epoch of 45. The initial learning rates is set to 1$\times$$10^{-5}$ and 3$\times$$10^{-4}$ for the backbone layers and new added layers, respectively. We adjust learning rate with a step decay schedule. The decay factor is set to 0.1 and associated with a decay epoch of 15.

\subsection{Evaluation metrics}
We consider accuracy and two semantic measures for assessing performance of models:
\noindent\textbf{Mistake severity (MS):} Inspired by \cite{bertinetto2020making}, we measure the severity of a mis-classification with the height of the LCA between the class of incorrect prediction and the true class. \textbf{Hierarchical distance (HD) of of top-}\textit{k}\footnote{We set k as 5 in this study.}\textbf{:} This metric computes the mean hierarchical class distance between the true class and the top-\textit{k} predicted classes. The measurement is meaningful for assessing the reliance of a model in assisting clinicians making diagnosis.

\subsection{Quantitative Results}
\subsubsection{Ablation study:} 
\begin{figure}[t!]
\centering
\includegraphics[width=0.80\textwidth]{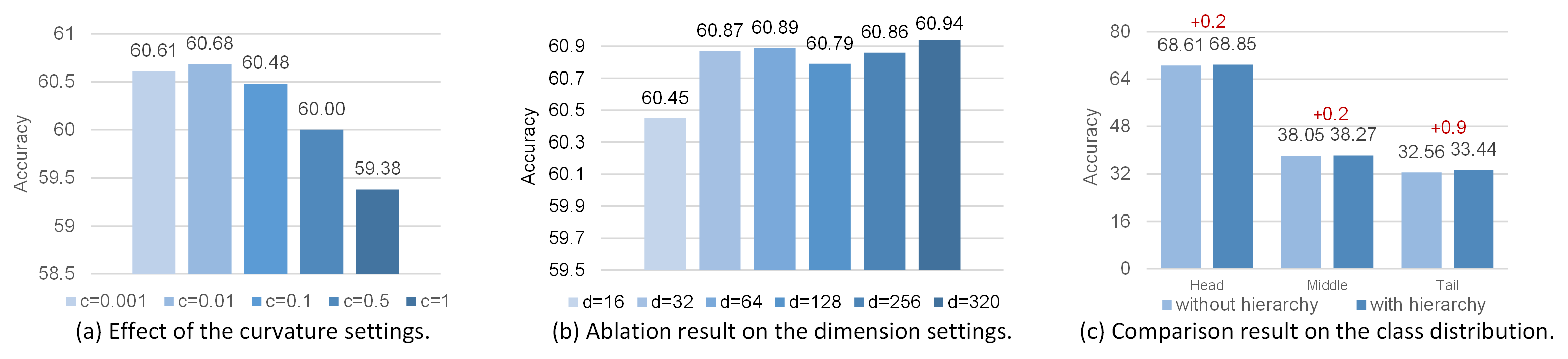}
\caption{Ablation results on the curvature setting and the dimension setting.}
\label{fig: ablation}
\end{figure}

Here, we give ablation results of our model to illustrate how different settings affect the final performance. Fig. \ref{fig: ablation}~(a) shows the effect of curvature which determines the distortion of the hyperbolic ball. It can be seen that a small \textit{c} produces better performance and the accuracy drops 1.6\% when increasing \textit{c} from 0.01 to 1. As demonstrated by \cite{khrulkov2020hyperbolic}, this is because large curvatures could bring numerical instability in hyperbolic operations. Therefore, we set \textit{c} as 0.01 in the following experiments. Then, we report accuracy by varying the hyperbolic dimension in Fig. \ref{fig: ablation}~(b). It can be noted that the model with dimension of 320 achieves highest accuracy of 60.94\%. However, the performance gap is not significant compared to other models with lower dimension settings. Even reducing the dimension from 320 to 16, the accuracy only decreases $\sim$0.5\%. This result verifies the efficiency of hyperbolic embeddings in representing imaging data. In Table \ref{tab:cmp_result}, we compare the performance of the HPN trained with and without class hierarchy. It can be seen that both the HCD and LCD encoding boost the performance compared to the baseline hyperbolic network. Among them, the HPN trained with the class relation matrix derived from LCA encoding gives best accuracy and semantic metrics. Since our dataset is highly imbalanced, in Fig. \ref{fig: ablation}~(c), we further give the detailed performance improvement on head classes, middle class and tail classes separately. When using the class hierarchy, the accuracy increases 0.9\% for tail classes which is higher than that of 0.2\% for both middle and head classes.

\begin{table}[!t]
\scriptsize
\centering
\caption{Comparison of the proposed model with other methods.}
\begin{tabular}{c|c|c|cc}
\hline\hline
\multirow{2}{*}{Models} & \multirow{2}{*}{Class hierarchy} & \multirow{2}{*}{Accuracy} & \multicolumn{2}{c}{Semantic Metrics} \\ \cline{4-5} 
 &  &  & \multicolumn{1}{c|}{Mistake severity} & Mean HD@5 \\ \hline\hline
Baseline CNN & \xmark & 58.47 & \multicolumn{1}{c|}{1.95} & 1.65 \\ \hline
Soft-label\cite{bertinetto2020making} & \cmark & 58.88 & \multicolumn{1}{c|}{1.95} & 1.46 \\ \hline
Multi-branch CNN \cite{zhu2017b} & \cmark & 60.28 & \multicolumn{1}{c|}{1.92} & 1.61\\ \hline
\begin{tabular}[c]{@{}c@{}} Fixed-hyperbolic  \\ embeddings\end{tabular} \cite{long2020searching} & \cmark & 56.87 & \multicolumn{1}{c|}{1.85} & 1.40 \\ \hline
\begin{tabular}[c]{@{}c@{}} Euclidean  \\ prototype net \end{tabular}\cite{garnot2021leveraging} & \cmark~(LCD) & 60.76 & \multicolumn{1}{c|}{1.90} & 1.55 \\ \hline
\multirow{3}{*}{\begin{tabular}[c]{@{}c@{}}Hyperbolic \\ prototype net\end{tabular}} & \xmark  & 60.68 & \multicolumn{1}{c|}{1.91} & 1.63 \\ 
\cline{2-5} &  \cmark~(HCD) & \multicolumn{1}{c|}{60.94} & \multicolumn{1}{c|}{1.83} &\multicolumn{1}{c}{1.43} \\ 
\cline{2-5} &  \cmark~(LCD) & \multicolumn{1}{c|}{61.04} & \multicolumn{1}{c|}{1.82} &\multicolumn{1}{c}{1.41} \\ 
\hline\hline
\end{tabular}\label{tab:cmp_result}
\end{table}

\begin{figure}[!t]
\centering
\includegraphics[width=0.90\textwidth]{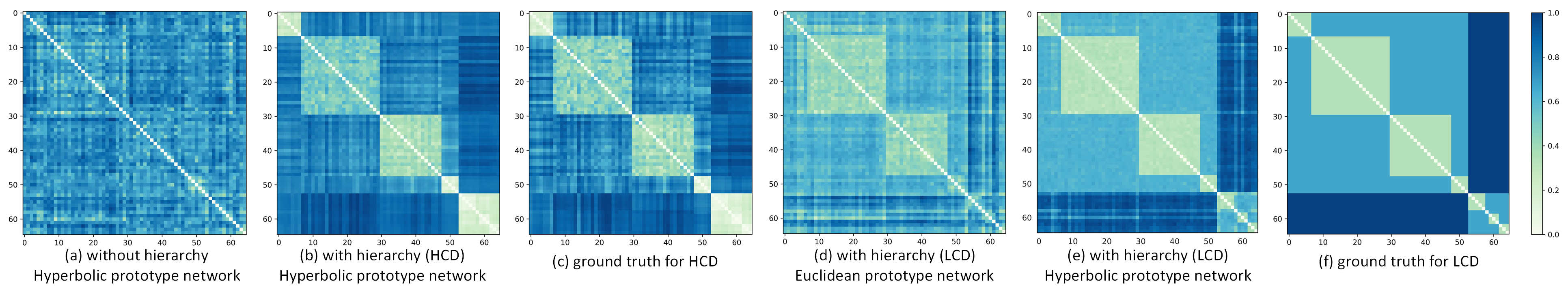}
\caption{Pairwise class prototype distance learned with and without class hierarchy.}
\label{fig: pred_class_distance}
\end{figure}

\subsubsection{Comparative study:} We then compare our model with that of models trained with and without class hierarchy in Euclidean space and hyperbolic space, respectively. The details of those model are described in the Appendix. From Table \ref{tab:cmp_result}, we can observe that all class-hierarchy trained models apart from the method with fixed-hyperbolic embeddings show higher accuracy compared to the baseline CNN. While the semantic metric of all models learned with class hierarchy are better than that of the baseline CNN. Certainly, this result highlights the value of incorporating class relations for skin lesion recognition. For the fixed-hyperbolic embeddings that achieves best semantic measurements with the lowest accuracy, we attribute this discrepancy to the softmax-based cross-entropy optimization. Because it is hard to minimize the loss on the probability distribution of distance~(eq.(\ref{eq:dist_softmax})) between an image representation and fixed prototypes with a semantic distance constraint. Noticeably, when using class hierarchy, our model outperforms the Euclidean prototype net which has best performance among all Euclidean models.

\begin{figure}[t!]
\centering
\includegraphics[width=0.99\textwidth]{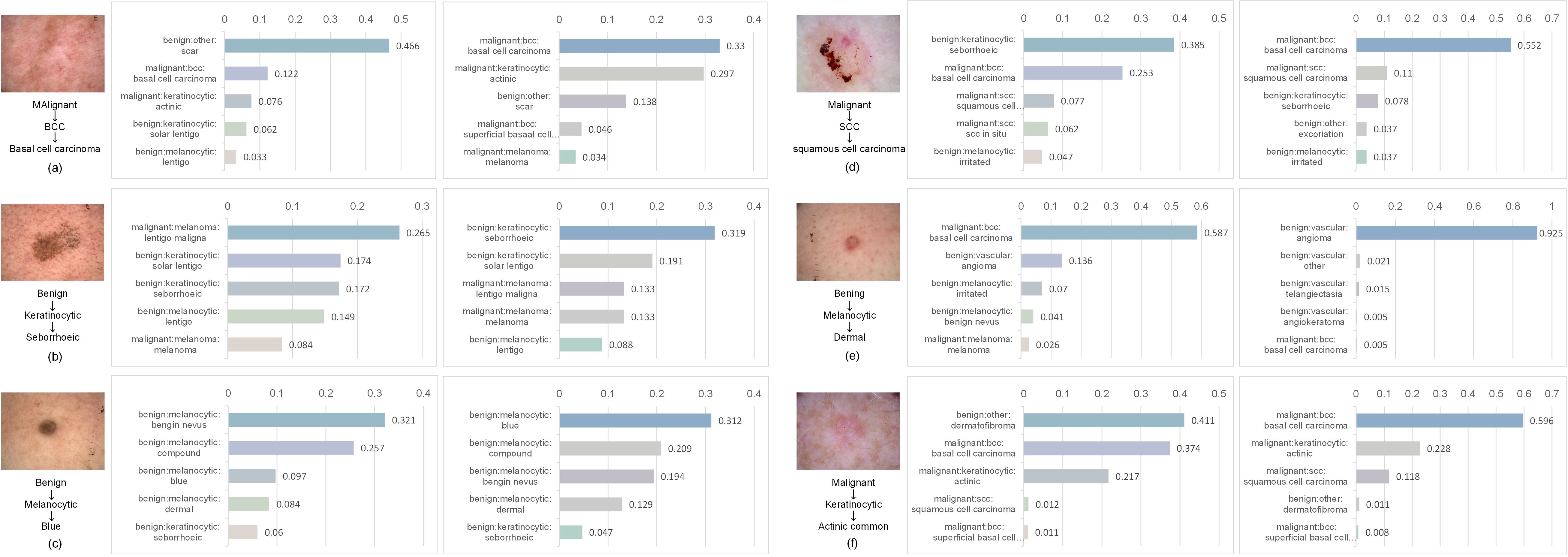}
\caption{Examples of predictions results from our model trained with and without the class hierarchy. (a)-(c) are incorrectly predicted samples by both model. (d)-(f) are samples correctly predicted by hierarchy-aware HPN but mis-classified by hierarchy-agnostic HPN (best viewed in zoom in mode).}
\label{fig: pred_samples}
\end{figure}

\subsection{Visualization results}
In Fig. \ref{fig: pred_class_distance}, we illustrate the distance matrix for class prototypes learned by our hyperbolic network and the Euclidean prototype network. It can be note that there is no clear relation for our model trained without using hierarchical class clues.  By contrast, both the class-hierarchy regularized models shows a semantic connection in the prototypical class distance matrix. From Fig.~\ref{fig: pred_class_distance}(d) and Fig.~\ref{fig: pred_class_distance}(e), we can see the pairwise prototypes distance of the hyperbolic network is closer to the ground truth class distance compare to that of Euclidean model. We then give prediction results for individual samples in Fig.~\ref{fig: pred_samples}. It can be seen the hierarchy-aware HPN gives more reasonable predictions compared with the hierarchy-agnostic HPN. 

\section{Conclusion}
In this study, we present a hyperbolic network with class hierarchy for skin lesion recognition. Our model is capable of jointly learning hyperbolic image embeddings and class prototypes while preserve class relations from the hierarchy. We evaluate the proposed method on a large-scale in-house skin lesion dataset by reporting both accuracy and semantic measurements derived from the class tree. Experiments demonstrate that our model can capture well class relations and the hyperbolic network outperforms other Euclidean models.

\bibliographystyle{splncs04.bst}
\bibliography{main.bib}

\begin{thebibliography}{10}
\providecommand{\url}[1]{\texttt{#1}}
\providecommand{\urlprefix}{URL }
\providecommand{\doi}[1]{https://doi.org/#1}

\bibitem{barata2019deep}
Barata, C., Marques, J.S., Celebi, M.E.: Deep attention model for the
  hierarchical diagnosis of skin lesions. In: 2019 IEEE/CVF Conference on
  Computer Vision and Pattern Recognition Workshops. pp. 2757--2765 (2019)

\bibitem{bertinetto2020making}
Bertinetto, L., Mueller, R., Tertikas, K., Samangooei, S., Lord, N.A.: Making
  better mistakes: Leveraging class hierarchies with deep networks. In:
  Proceedings of the IEEE/CVF Conference on Computer Vision and Pattern
  Recognition. pp. 12506--12515 (2020)

\bibitem{brinker2019deep}
Brinker, T.J., Hekler, A., Enk, A.H., Klode, J., Hauschild, A., Berking, C.,
  Schilling, B., Haferkamp, S., Schadendorf, D., Holland-Letz, T., et~al.: Deep
  learning outperformed 136 of 157 dermatologists in a head-to-head dermoscopic
  melanoma image classification task. European Journal of Cancer  \textbf{113},
   47--54 (2019)

\bibitem{esteva2017dermatologist}
Esteva, A., Kuprel, B., Novoa, R.A., Ko, J., Swetter, S.M., Blau, H.M., Thrun,
  S.: Dermatologist-level classification of skin cancer with deep neural
  networks. nature  \textbf{542}(7639),  115--118 (2017)

\bibitem{ganea2018hyperbolic}
Ganea, O., B{\'e}cigneul, G., Hofmann, T.: Hyperbolic entailment cones for
  learning hierarchical embeddings. In: International Conference on Machine
  Learning. pp. 1646--1655 (2018)

\bibitem{garnot2021leveraging}
Garnot, V.S.F., Landrieu, L.: Leveraging class hierarchies with metric-guided
  prototype learning  (2021)

\bibitem{ge2017skin}
Ge, Z., Demyanov, S., Chakravorty, R., Bowling, A., Garnavi, R.: Skin disease
  recognition using deep saliency features and multimodal learning of
  dermoscopy and clinical images. In: International conference on medical image
  computing and computer-assisted intervention. pp. 250--258. Springer (2017)

\bibitem{ijcai2021-337}
Goyal, P., Choudhary, D., Ghosh, S.: Hierarchical class-based curriculum loss.
  In: Proceedings of the Thirtieth International Joint Conference on Artificial
  Intelligence. pp. 2448--2454 (2021)

\bibitem{he2016deep}
He, K., Zhang, X., Ren, S., Sun, J.: Deep residual learning for image
  recognition. In: Proceedings of the IEEE conference on computer vision and
  pattern recognition. pp. 770--778 (2016)

\bibitem{khrulkov2020hyperbolic}
Khrulkov, V., Mirvakhabova, L., Ustinova, E., Oseledets, I., Lempitsky, V.:
  Hyperbolic image embeddings. In: Proceedings of the IEEE/CVF Conference on
  Computer Vision and Pattern Recognition. pp. 6418--6428 (2020)

\bibitem{long2020searching}
Long, T., Mettes, P., Shen, H.T., Snoek, C.G.: Searching for actions on the
  hyperbole. In: Proceedings of the IEEE/CVF Conference on Computer Vision and
  Pattern Recognition. pp. 1141--1150 (2020)

\bibitem{nickel2017poincare}
Nickel, M., Kiela, D.: Poincar{\'e} embeddings for learning hierarchical
  representations. Advances in neural information processing systems
  \textbf{30} (2017)

\bibitem{sala2018representation}
Sala, F., De~Sa, C., Gu, A., R{\'e}, C.: Representation tradeoffs for
  hyperbolic embeddings. In: International conference on machine learning. pp.
  4460--4469. PMLR (2018)

\bibitem{yang2020hierarchical}
Yang, J., Gao, M., Kuang, K., Ni, B., She, Y., Xie, D., Chen, C.: Hierarchical
  classification of pulmonary lesions: A large-scale radio-pathomics study. In:
  International Conference on Medical Image Computing and Computer-Assisted
  Intervention. pp. 497--507 (2020)

\bibitem{yu2016automated}
Yu, L., Chen, H., Dou, Q., Qin, J., Heng, P.A.: Automated melanoma recognition
  in dermoscopy images via very deep residual networks. IEEE transactions on
  medical imaging  \textbf{36}(4),  994--1004 (2016)

\bibitem{zhu2017b}
Zhu, X., Bain, M.: B-cnn: branch convolutional neural network for hierarchical
  classification. arXiv preprint arXiv:1709.09890  (2017)

\end{thebibliography}



\end{document}